\documentclass[runningheads]{llncs}
\usepackage[T1]{fontenc}
\usepackage{graphicx}
\usepackage{booktabs}
\usepackage[misc]{ifsym}
\newcommand{\corr}{(\Letter)}
\usepackage{url}
\usepackage[section]{placeins}

\begin{document}

\title{EssayCBM: Transparent Essay Scoring via Rubric-Aligned Concept Bottlenecks}

\titlerunning{Concept Bottleneck Models for Transparent Essay Scoring}

\author{
Kumar Satvik Chaudhary \corr \and
Chengshuai Zhao \and
Fan Zhang \and
Garima Agrawal \and
Yuli Deng \and
Huan Liu
}

\authorrunning{K. S. Chaudhary et al.}

\institute{
Arizona State University, Tempe, AZ, USA\\
\email{\{kchaud13, czhao93, fzhan113, garima.agrawal, ydeng19, huanliu\}@asu.edu}
}

\toctitle{EssayCBM: Transparent Essay Scoring via Rubric-Aligned Concept Bottlenecks}
\tocauthor{Kumar Satvik Chaudhary, Chengshuai Zhao, Fan Zhang, Garima Agrawal, Yuli Deng, Huan Liu}

\maketitle

\begin{abstract}
Automated essay scoring (AES) has advanced with neural language models, but many systems still provide limited visibility into how grades are produced. In educational settings, instructors often use structured rubrics and need to inspect intermediate judgments before accepting a final score. We introduce EssayCBM, a rubric-aligned concept bottleneck framework that decomposes essay evaluation into eight writing concepts and computes the final grade through an auditable concept-to-grade mapping. The demo lets users view concept predictions, examine rubric-level confidence, adjust concept scores, and update the final grade without re-encoding the essay. Experiments show that EssayCBM remains competitive with neural AES baselines while supporting a more controllable and transparent grading workflow. Code and demo are available at \url{https://github.com/scott-f-zhang/CBM-Demo} and \url{https://youtu.be/tkIGJTN4ZVU}.

\end{abstract}

\section{Introduction}

Automated essay scoring (AES) has advanced significantly with neural language models \cite{alikaniotis2016automatic,taghipour2016aes}. Modern systems can approximate human grading performance by learning representations of writing quality directly from text. However, most AES models remain opaque, predicting a final score without exposing the intermediate judgments behind it. This lack of transparency is problematic in educational settings, where instructors evaluate essays using structured rubrics that assess dimensions such as thesis clarity, organization, and evidence use. When these dimensions are collapsed into a single prediction, instructors cannot verify or correct the rubric-level judgments behind the final grade.

\begin{figure}[t]
\centering
\includegraphics[width=0.60\linewidth]{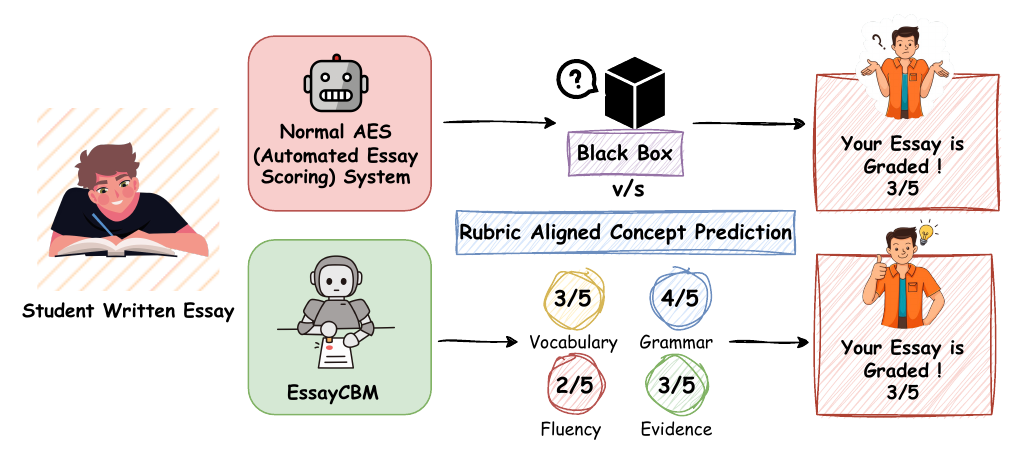}
\caption{Transparent concept-level grading in EssayCBM versus black-box AES systems}
\label{fig:architecture}
\end{figure}

Recent work has shown the effectiveness of large language models in AI-assisted education~\cite{zhao2025cyberbot}. However, when applied to automated grading, LLM-based approaches often rely on prompt-based scoring, where the relationship between rubric-level judgments and the final grade remains implicit. As a result, scores may vary across runs, and instructors have limited ability to audit or intervene on the intermediate scoring process.

To address these challenges, we introduce EssayCBM, a rubric-aligned concept bottleneck framework \cite{koh2020cbm} for interpretable essay scoring. EssayCBM models eight writing concepts corresponding to rubric dimensions such as thesis clarity, evidence use, and organization. The system first predicts scores for these concepts, and the final grade is computed strictly as a function of these intermediate predictions. By constraining grading decisions to pass through explicit concept variables, EssayCBM provides rubric-level interpretability through a transparent and traceable concept-to-grade mapping. While many AES dashboards mainly present scores or feedback, our demo lets instructors inspect and revise rubric-level concept predictions and update the final grade during grading.

\textbf{Contributions.}
(1) We present EssayCBM, a concept bottleneck framework that enables rubric-level interpretability in automated essay scoring.
(2) We show that EssayCBM matches neural AES baselines with a stable, auditable concept-to-grade mapping.
(3) We present an interactive interface for inspecting concept predictions, adjusting rubric scores, and updating grades in real time.

\section{EssayCBM Method}

Standard AES models learn a direct mapping $f:X\rightarrow Y$, where essay $X$ is mapped to final grade $Y$ from neural text representations. EssayCBM instead uses a concept bottleneck formulation, constraining prediction to pass through an explicit rubric-level concept layer: $X \rightarrow C \rightarrow Y$. Unlike prompt-based LLM grading, EssayCBM enforces a structural constraint in which the final grade is computed solely from inspectable concept predictions.

\begin{figure}[!t]
\centering
\includegraphics[width=0.96\linewidth]{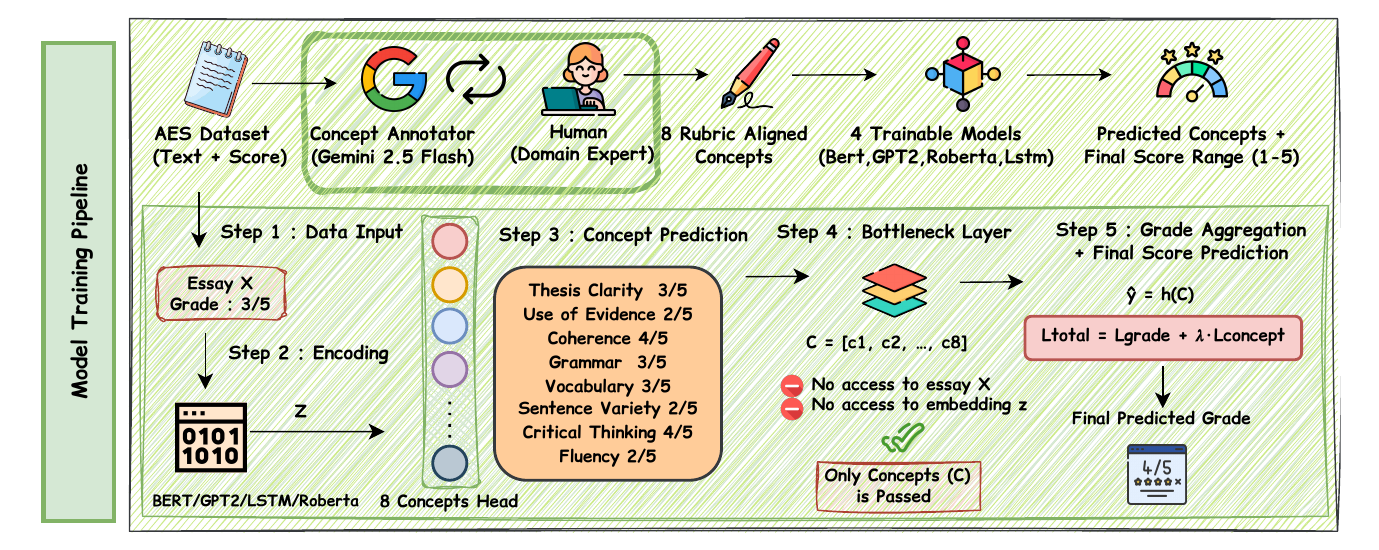}
\caption{Overview of the EssayCBM framework}
\label{fig:concept}
\end{figure}

\textbf{Concept Prediction ($X\rightarrow C$).}
Given essay $X$, a pretrained encoder produces representation $z=f_{\theta}(X)$. Eight classifier heads predict rubric-aligned concept scores $c_k=g_k(z),\,k=1,\ldots,8$, forming $C=\{c_1,\ldots,c_8\}$. Each concept corresponds to a rubric dimension used in essay evaluation: Thesis Clarity, Evidence Use, Organization and Coherence, Grammar, Vocabulary, Sentence Variety, Critical Thinking, and Fluency. Concept labels were generated using Gemini 2.5 Flash and verified for consistency.

 \textbf{Concept Bottleneck ($C\rightarrow Y$).}
The concept vector $C$ forms the bottleneck representation used to compute the predicted grade $\hat{Y}=h(C)$. Model parameters are trained using the objective $\mathcal{L}_{total}=\mathcal{L}_{grade}+\lambda\mathcal{L}_{concept}$, jointly optimizing grade prediction and concept prediction accuracy. Because $h$ receives only $C$, grading decisions depend exclusively on rubric-level concept scores.

\textbf{Human-in-the-Loop Intervention.}
The final grade depends only on concept scores, allowing instructors to revise any incorrect $c_k$. The system then recomputes $\hat{Y}=h(C')$ from the updated concept vector without re-running the encoder, enabling grade correction while preserving interpretability.

\section{Demo and Evaluation}

EssayCBM is an interactive web demo where users submit an essay and receive a predicted grade, rubric-level concept scores, and confidence plots. Instructors can revise inaccurate concept scores and view the updated grade. The system uses a Streamlit frontend and FastAPI backend for model loading, inference, database access, and routing.

\begin{table}[!h]
\centering
\caption{Baseline vs. EssayCBM performance (accuracy / macro-F1, in \%).}
\label{tab:encoder-results}
\begin{tabular}{l c @{\hspace{10pt}} c}
\toprule
Models & Baseline & EssayCBM \\
\midrule
BERT    & 80.70 / 60.01 & 81.14 / 62.38 \\
RoBERTa & 80.70 / 61.98 & 79.39 / 58.88 \\
GPT-2   & 78.07 / 57.81 & 79.03 / 56.21 \\
LSTM    & 79.39 / 44.25 & 80.19 / 44.21 \\
\bottomrule
\end{tabular}
\end{table}

We evaluate EssayCBM on the Automated Essay Scoring benchmark using BERT, RoBERTa, GPT-2, and LSTM encoders. As shown in Table~\ref{tab:encoder-results}, EssayCBM remains competitive with direct AES baselines, showing that rubric-level interpretability can be introduced without substantially sacrificing performance.

\begin{figure}[!t]
\centering
\includegraphics[width=0.80\linewidth]{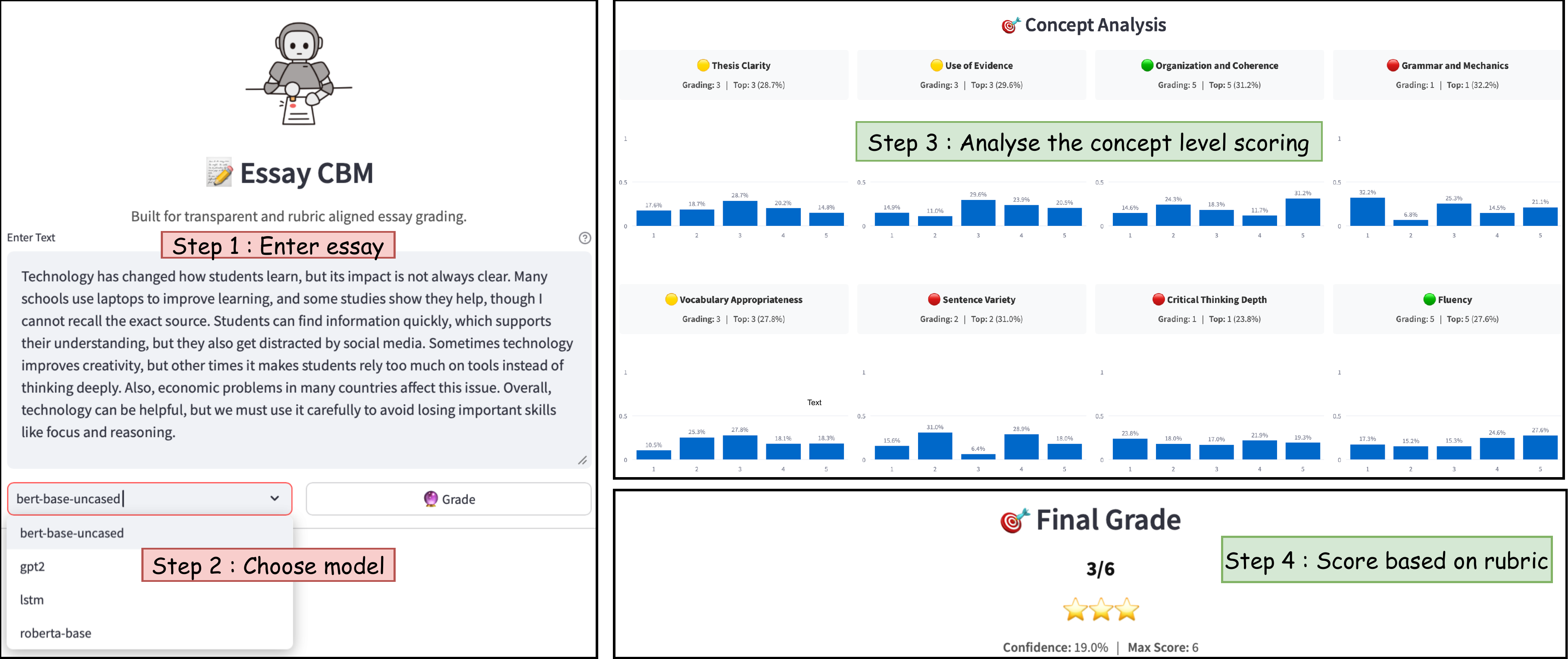}
\caption{EssayCBM Streamlit interface showing concept scores and final grade}
\label{fig:demo}
\end{figure}

\textbf{Limitations.}
Explanation reliability depends on concept prediction accuracy, as incorrect scores may produce misleading explanations. Concept annotation is also labor- and cost-intensive. The current concept set targets argumentative essays and requires adaptation for other genres. Future work should assess agreement between instructor-provided and system-predicted scores.

\section{Conclusion}

EssayCBM presents a rubric-aligned concept bottleneck framework for transparent essay scoring. Its editable concept-level judgments allow instructors to adjust rubric scores and update the final grade. The demo illustrates how concept-based grading supports controllable human-AI collaboration in automated assessment.

\section*{Acknowledgments.}
This work is supported by the National Science Foundation (NSF) under grants  SaTC (\#2335666) and IIS-2229461.

\end{document}